\documentclass[11pt]{article}

\usepackage[margin=1in]{geometry}
\usepackage[numbers]{natbib}

\usepackage{times}
\usepackage{hyperref}
\usepackage{url}
\usepackage{booktabs}
\usepackage{amsfonts}
\usepackage{nicefrac}
\usepackage{microtype}
\usepackage{xcolor}
\usepackage{amsmath}
\usepackage{amssymb}
\usepackage{listings}
\usepackage{algorithm}
\usepackage{algorithmic}
\usepackage{tabularx}
\usepackage{multirow}
\usepackage{adjustbox}
\usepackage{graphicx}

\title{CosmicFish-HRM: Adaptive Reasoning via Hierarchical Recurrent Mechanisms in Compact Language Models}

\author{
    Venkat Akhil Lakkapragada\\
    Mistyoz AI\\
    Hyderabad, India\\
    \texttt{lvakhil06@gmail.com}
}

\begin{document}

\maketitle

\begin{abstract}
Large language models have achieved strong reasoning capabilities, though often at the cost of massive parameter counts and expensive inference. In this work, we explore a different direction: adaptive reasoning depth in compact language models.

We present CosmicFish-HRM, a compact language model built around a Hierarchical Reasoning Module (HRM) that dynamically allocates reasoning compute during inference. Instead of applying fixed computation to every input, the model iterates through high-level and low-level reasoning cycles and learns when to halt based on input complexity.

CosmicFish-HRM combines this adaptive reasoning core with modern transformer components including Grouped Query Attention, RoPE, and SwiGLU activations. While the additional reasoning infrastructure introduces overhead at small scale, we hypothesize that this tradeoff becomes increasingly favorable as model size grows and the relative cost of the HRM core diminishes.

Our results show that the model learns non-uniform reasoning behavior, allocating different numbers of reasoning steps across tasks and inputs. These findings suggest that adaptive reasoning depth may offer a promising alternative to relying solely on parameter scale for reasoning capability.
\end{abstract}

\section{Introduction}

The dominant trend in modern language modeling has been scale. From GPT-4~\cite{openai2023gpt4} to LLaMA~\cite{touvron2023llama}, progress has largely followed a familiar pattern: larger parameter counts, larger datasets, and increasingly expensive training runs. The results have been impressive, especially in reasoning and instruction following, though this progress has also pushed state-of-the-art systems further away from practical deployment on edge hardware, mobile devices, and other resource-constrained environments. Running modern language models often requires substantial memory, specialized accelerators, and inference infrastructure that simply does not exist outside datacenters.

Most existing approaches to efficiency focus on compression. Quantization, pruning, and knowledge distillation~\cite{hinton2015distilling} reduce the size of already-large models, but the underlying computational structure usually remains unchanged. A standard transformer still applies the same depth of computation to every input regardless of difficulty. A short factual completion and a multi-step reasoning problem pass through essentially the same pipeline with the same number of layers and attention operations. In practice, though, not every prompt demands the same amount of reasoning.

We take a different view. Some tasks are almost reflexive. Others require iteration, revision, or longer chains of internal computation. A compact model capable of allocating additional reasoning steps only when necessary may offer a different tradeoff from simply increasing parameter count indefinitely. The central idea behind this work is straightforward: reasoning ability may depend not only on model scale, but also on how flexibly computation is used during inference.

To explore this idea, we introduce \textbf{CosmicFish-HRM}, a compact 82.77M parameter language model built around a \textit{Hierarchical Reasoning Module} (HRM). Rather than operating as a purely fixed-depth transformer, the model alternates between high-level ($H$) and low-level ($L$) reasoning states through iterative recurrent cycles. A learned halting head determines whether additional reasoning steps are necessary before generation continues. Difficult inputs may trigger deeper reasoning trajectories, while simpler prompts can halt early and avoid unnecessary computation.

The architecture combines this adaptive reasoning core with modern transformer components including Rotary Positional Embeddings (RoPE)~\cite{su2021roformer}, Grouped Query Attention (GQA)~\cite{ainslie2023gqa}, and SwiGLU activations~\cite{shazeer2020glu}. CosmicFish-HRM was trained on \textit{CosmicSet}, a curated 10B token dataset spanning web text, Wikipedia, code, mathematics, and research papers.

At compact scales, introducing recurrent reasoning infrastructure creates a tradeoff. Part of the model capacity is allocated to adaptive reasoning behavior rather than raw language modeling capacity alone. This can limit benchmark performance relative to conventional transformers of similar size. We hypothesize, however, that this tradeoff becomes increasingly favorable as model scale grows and the relative overhead of the reasoning infrastructure diminishes while the adaptive computation behavior remains intact.

This paper makes the following contributions:

\begin{itemize}
    \item We propose \textbf{CosmicFish-HRM}, a compact language model architecture that integrates a Hierarchical Reasoning Module within a transformer backbone to enable adaptive reasoning depth during inference.

    \item We introduce a learned adaptive halting mechanism that allows the model to dynamically vary reasoning depth across inputs instead of applying fixed computation uniformly.

    \item We analyze the relationship between reasoning step allocation and input complexity, showing that the model exhibits non-uniform reasoning behavior across tasks.

    \item We discuss the tradeoffs introduced by adaptive reasoning infrastructure at compact scale and argue that these costs may amortize more favorably as model size increases.
\end{itemize}

The remainder of this paper is organized as follows. Section~\ref{sec:related} reviews related work on adaptive computation and efficient language models. Section~\ref{sec:architecture} describes the CosmicFish-HRM architecture in detail. Section~\ref{sec:experiments} presents experimental results and evaluation. Section~\ref{sec:analysis} analyzes the model's adaptive reasoning behavior, and Section~\ref{sec:conclusion} discusses limitations and future directions.

\section{Related Work}
\label{sec:related}

\subsection{Adaptive Computation in Neural Networks}

The idea that neural networks should adapt their computational effort based on input difficulty is not new. Graves~\cite{graves2016adaptive} introduced Adaptive Computation Time (ACT), which allowed recurrent networks to decide how many computational steps to use before producing an output. The core intuition was simple but important: different inputs may require different amounts of processing. ACT, however, was developed in the context of recurrent networks and does not naturally extend to modern transformer-based language models.

Later work explored related forms of adaptive computation. PonderNet~\cite{banino2021pondernet} introduced a probabilistic halting mechanism trained with an additional regularization objective, encouraging models to "ponder" only as long as necessary. Universal Transformers~\cite{dehghani2018universal} incorporated recurrent depth into transformer architectures by repeatedly applying the same layer across multiple iterations, though without an explicit learned halting policy. Confident Adaptive Language Modeling~\cite{schuster2022confident} explored early exiting in transformer decoders, allowing inference to stop at intermediate layers once prediction confidence became sufficiently high.

Mixture of Depths~\cite{raposo2024mixture} approached the problem differently by routing tokens dynamically through subsets of transformer layers. Rather than iterative reasoning, this approach focuses on allocating computation unevenly across tokens and layers to reduce overall inference cost.

CosmicFish-HRM is motivated by a somewhat different question. Instead of early exits or token routing alone, we investigate whether compact language models can benefit from iterative reasoning cycles operating across multiple abstraction levels before generation proceeds.

\subsection{Hierarchical Reasoning Models}

The work most directly connected to ours is the Hierarchical Reasoning Model (HRM) introduced by Wang et al.~\cite{wang2025hrm}. HRM proposed a recurrent reasoning architecture composed of two interacting modules: a slower high-level reasoning system responsible for abstract planning, and a faster low-level system responsible for detailed computation. Despite its relatively small parameter count, the original HRM achieved strong performance on structured reasoning tasks including Sudoku solving, maze navigation, and ARC-style reasoning benchmarks~\cite{chollet2019arc}.

That said, the original HRM was not designed as a language model. It operated primarily as a task-specific reasoning engine trained on relatively small puzzle-oriented datasets, without open-ended text generation or autoregressive language modeling capabilities.

CosmicFish-HRM builds on the central intuition behind HRM while moving into a different setting. Rather than treating hierarchical reasoning as a standalone reasoning engine, we integrate it into a decoder-only language model architecture trained on diverse natural language data. Our goal is not simply to reproduce HRM-style reasoning behavior, but to explore whether iterative hierarchical computation can provide useful adaptive reasoning dynamics inside compact autoregressive language models.

\subsection{Small and Efficient Language Models}

A separate line of work has focused on improving the capabilities of small language models through training efficiency and data quality. TinyLlama~\cite{zhang2024tinyllama} demonstrated that relatively small models trained on large-scale corpora could remain competitive despite reduced parameter count. The Phi series~\cite{gunasekar2023textbooks} further argued that carefully curated high-quality datasets can significantly improve compact model performance.

These approaches primarily improve efficiency through data and training strategy rather than changes to inference-time computation itself. The underlying architecture generally remains a conventional fixed-depth transformer.

CosmicFish-HRM explores a complementary direction. Instead of relying solely on improved data efficiency, the model attempts to vary reasoning depth dynamically during inference. At compact scales, this introduces an architectural tradeoff: some fraction of model capacity is devoted to adaptive reasoning infrastructure rather than raw language modeling capacity alone. We view this work as an exploration of that tradeoff rather than a claim that adaptive reasoning universally improves small-model benchmark performance.

\subsection{Adaptive Halting and Learned Decision Mechanisms}

Neural networks have previously incorporated learned decision mechanisms for routing, halting, and adaptive execution~\cite{mnih2015dqn}. Inspired by these ideas, CosmicFish-HRM uses a lightweight learned halting head to determine whether additional reasoning iterations are necessary at each step.

Our formulation is intentionally simple. The halting mechanism is trained jointly with the language modeling objective and step penalty, allowing the model to learn variable-depth reasoning behavior without requiring manually designed stopping criteria. Rather than framing the system as a full reinforcement learning setup, we treat the halting behavior primarily as a learned adaptive computation mechanism integrated directly into the forward pass.

\section{Architecture}
\label{sec:architecture}

CosmicFish-HRM is a decoder-only language model organized around three major stages: an input transformer stack for contextual representation learning, a Hierarchical Reasoning Module (HRM) for adaptive iterative computation, and an output transformer stack for token prediction. The overall design separates linguistic representation learning from the adaptive reasoning process itself.

\begin{figure}[h]
    \centering
    \includegraphics[width=0.85\linewidth]{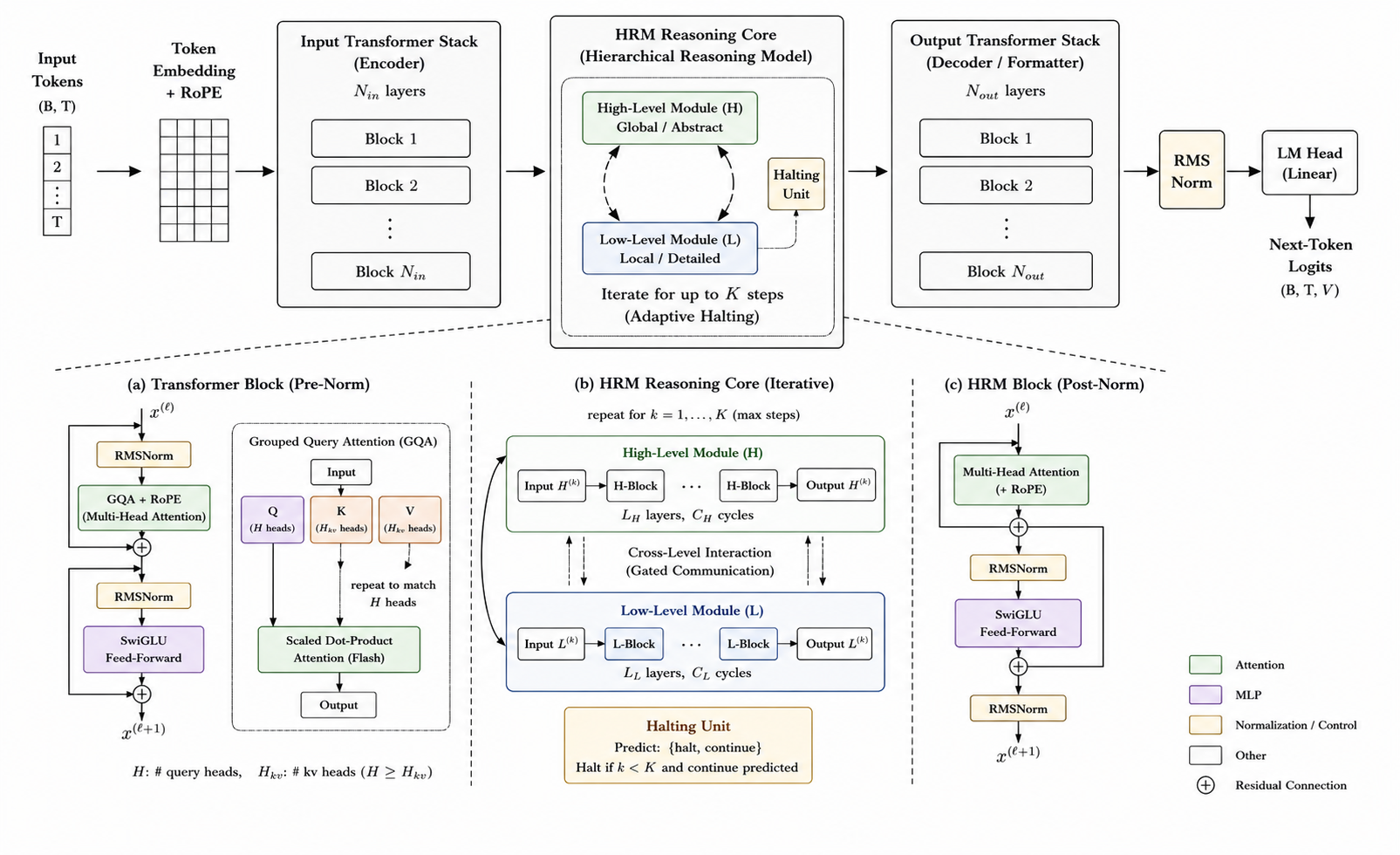}
    \caption{Overview of the CosmicFish-HRM architecture. Input tokens are embedded and processed by transformer blocks, passed through the HRM reasoning core for a variable number of reasoning iterations, and decoded into next-token predictions through output transformer layers.}
    \label{fig:architecture}
\end{figure}

\subsection{Overall Design}
\label{sec:arch_overall}

The architecture follows the pipeline:

\begin{equation}
    \mathbf{x} \xrightarrow{\text{Embed}}
    \mathbf{h}_0 \xrightarrow{\text{Input Blocks}}
    \mathbf{h} \xrightarrow{\text{HRM Core}}
    \mathbf{z} \xrightarrow{\text{Output Blocks}}
    \hat{\mathbf{z}} \xrightarrow{\text{LM Head}}
    \mathbf{logits}
\end{equation}

Given an input token sequence
$\mathbf{x} = (x_1, x_2, \ldots, x_T)$, tokens are first mapped into dense embeddings through a learned embedding matrix $\mathbf{E} \in \mathbb{R}^{V \times d}$, where $V = 50304$ is the vocabulary size and $d = 448$ is the embedding dimension. The representation is then processed through six input transformer layers, the HRM reasoning core, and six output transformer layers before projection into vocabulary logits. Following standard practice, the embedding matrix is weight-tied with the language modeling head~\cite{press2017tying}.

Key architectural hyperparameters are summarized in Table~\ref{tab:hyperparams}.

\begin{table}[h]
\centering
\caption{CosmicFish-HRM architecture hyperparameters.}
\label{tab:hyperparams}
\begin{tabular}{lc}
\toprule
\textbf{Hyperparameter} & \textbf{Value} \\
\midrule
Total parameters & 82.77M \\
Embedding dimension $d$ & 448 \\
Vocabulary size $V$ & 50,304 \\
Maximum sequence length & 512 \\
Input transformer layers & 6 \\
Output transformer layers & 6 \\
Attention heads & 8 \\
KV heads (GQA) & 4 \\
HRM high-level layers $n_H$ & 4 \\
HRM low-level layers $n_L$ & 4 \\
Maximum reasoning steps & 16 \\
Exploration probability & 0.1 \\
Dropout & 0.1 \\
\bottomrule
\end{tabular}
\end{table}

\subsection{Input and Output Transformer Blocks}
\label{sec:arch_transformer}

Both transformer stacks follow a standard pre-normalization transformer design~\cite{xiong2020layer}. Each block applies attention and feedforward updates through residual connections:

\begin{equation}
    \mathbf{h}' = \mathbf{h} + \text{Attn}(\text{RMSNorm}(\mathbf{h}))
\end{equation}

\begin{equation}
    \mathbf{h}'' = \mathbf{h}' + \text{MLP}(\text{RMSNorm}(\mathbf{h}'))
\end{equation}

RMSNorm~\cite{zhang2019root} is used instead of LayerNorm for computational efficiency:

\begin{equation}
    \text{RMSNorm}(\mathbf{x}) =
    \frac{\mathbf{x}}{\text{RMS}(\mathbf{x})}
    \cdot \boldsymbol{\gamma}
\end{equation}

\begin{equation}
    \text{RMS}(\mathbf{x}) =
    \sqrt{\frac{1}{d}\sum_{i=1}^{d} x_i^2 + \epsilon}
\end{equation}

\paragraph{Grouped Query Attention.}

The model uses Grouped Query Attention (GQA)~\cite{ainslie2023gqa} with $n_q = 8$ query heads and $n_{kv} = 4$ key-value heads. Each query head has dimension $d_h = d / n_q = 56$. Sharing key-value heads across query groups reduces KV-cache memory requirements while preserving most of the representational capacity of standard multi-head attention.

Rotary Positional Embeddings (RoPE)~\cite{su2021roformer} are applied to queries and keys prior to attention computation:

\begin{equation}
    \mathbf{q}_m = \mathbf{R}_m \mathbf{q}_m,
    \quad
    \mathbf{k}_n = \mathbf{R}_n \mathbf{k}_n
\end{equation}

where $\mathbf{R}_m$ denotes the positional rotation matrix at position $m$. RoPE allows relative positional information to emerge directly through the attention mechanism without learned absolute position embeddings.

\paragraph{SwiGLU Feedforward Layers.}

Feedforward layers use the SwiGLU activation~\cite{shazeer2020glu}:

\begin{equation}
    \text{MLP}(\mathbf{x}) =
    \text{Dropout}
    \left(
    \mathbf{W}_{\text{down}}
    \left(
    \text{SiLU}(\mathbf{W}_{\text{up}}\mathbf{x})
    \odot
    \mathbf{W}_{\text{gate}}\mathbf{x}
    \right)
    \right)
\end{equation}

where
$\mathbf{W}_{\text{up}}, \mathbf{W}_{\text{gate}} \in \mathbb{R}^{4d \times d}$,
$\mathbf{W}_{\text{down}} \in \mathbb{R}^{d \times 4d}$,
and $\odot$ denotes elementwise multiplication.

\subsection{HRM Reasoning Core}
\label{sec:arch_hrm}

The HRM reasoning core forms the central architectural component of CosmicFish-HRM. Inspired by the Hierarchical Reasoning Model proposed by Wang et al.~\cite{wang2025hrm}, the system maintains two interacting recurrent reasoning states operating at different abstraction levels.

The high-level module $\mathcal{H}$ is intended to capture slower and more abstract reasoning behavior, while the low-level module $\mathcal{L}$ focuses on finer-grained local computation. Rather than replacing transformer layers entirely, the HRM core operates between transformer stacks as an adaptive reasoning stage.

\paragraph{State Initialization.}

Let $\mathbf{h} \in \mathbb{R}^{B \times T \times d}$ denote the output of the input transformer stack, where $B$ is the batch size and $T$ is the sequence length. The HRM initializes two recurrent states:

\begin{equation}
    \mathbf{z}_H^{(0)} = \mathbf{h},
    \quad
    \mathbf{z}_L^{(0)} = \mathbf{h}
\end{equation}

Both states begin from the same contextual representation and evolve independently during reasoning iterations.

\paragraph{Hierarchical Cycling.}

At each reasoning step $s \in \{1, \ldots, S\}$, the low-level module performs $c_L = 2$ recurrent cycles conditioned on the current high-level state:

\begin{equation}
    \mathbf{z}_L^{(s,k)} =
    \mathcal{L}
    \left(
    \mathbf{z}_L^{(s,k-1)},
    \mathbf{z}_H^{(s-1)}
    \right)
\end{equation}

where $k \in \{1, \ldots, c_L\}$ and $\mathcal{L}$ consists of $n_L = 4$ transformer blocks.

The updated low-level representation is then used to condition the high-level module:

\begin{equation}
    \mathbf{z}_H^{(s,k)} =
    \mathcal{H}
    \left(
    \mathbf{z}_H^{(s,k-1)},
    \mathbf{z}_L^{(s,c_L)}
    \right)
\end{equation}

where $\mathcal{H}$ consists of $n_H = 4$ transformer blocks and $k \in \{1, \ldots, c_H\}$ with $c_H = 2$.

This alternating update process creates a recurrent feedback loop between abstract and fine-grained reasoning states. Intuitively, lower-level representations refine details conditioned on higher-level context, while higher-level states update broader representations using information accumulated through lower-level computation.

\paragraph{Output Representation.}

After the halting mechanism terminates reasoning at step $S^*$, the final high-level representation $\mathbf{z}_H^{(S^*)}$ is passed into the output transformer stack for token prediction.

\subsection{Adaptive Halting Mechanism}
\label{sec:arch_halting}

A central component of CosmicFish-HRM is its learned adaptive halting mechanism. After each reasoning step $s$, a lightweight halting head computes halt and continue scores from the mean-pooled high-level representation:

\begin{equation}
    \mathbf{q}^{(s)} =
    \mathbf{W}_Q
    \left(
    \frac{1}{T}
    \sum_{t=1}^{T}
    \mathbf{z}_{H,t}^{(s)}
    \right)
    \in \mathbb{R}^2
\end{equation}

where $\mathbf{W}_Q \in \mathbb{R}^{2 \times d}$.

The model halts reasoning once the halt score exceeds the continue score:

\begin{equation}
    Q^{(s)}(\text{halt})
    >
    Q^{(s)}(\text{continue})
\end{equation}

\paragraph{Exploration During Training.}

During training, the model occasionally defers halting through an exploration mechanism designed to expose the network to longer reasoning trajectories:

\begin{equation}
    \text{halt}^{(s)} =
    \begin{cases}
    Q^{(s)}(\text{halt})
    >
    Q^{(s)}(\text{continue})
    & \text{with probability } 1 - p_{\text{explore}} \\
    \mathbb{1}
    [\text{steps} \geq \text{Uniform}(2, S_{\max})]
    & \text{with probability } p_{\text{explore}}
    \end{cases}
\end{equation}

where $p_{\text{explore}} = 0.1$ and $S_{\max} = 16$.

\paragraph{Inference Biasing.}

At inference time, a small halt bias $\delta = 0.35$ is added to encourage earlier stopping when additional reasoning cycles appear unnecessary:

\begin{equation}
    \hat{Q}^{(s)}(\text{halt}) =
    Q^{(s)}(\text{halt}) + \delta
\end{equation}

This creates a mild preference toward computational efficiency while still allowing deeper reasoning trajectories for more difficult inputs.

\subsection{Training Objective}
\label{sec:arch_objective}

CosmicFish-HRM is trained using a joint objective combining autoregressive language modeling with a lightweight reasoning-step penalty:

\begin{equation}
    \mathcal{L} =
    \mathcal{L}_{\text{LM}}
    +
    \lambda \cdot \bar{S}
\end{equation}

where $\mathcal{L}_{\text{LM}}$ is the standard next-token prediction loss:

\begin{equation}
    \mathcal{L}_{\text{LM}} =
    -\frac{1}{T}
    \sum_{t=1}^{T}
    \log
    P_\theta
    (x_{t+1} \mid x_1, \ldots, x_t)
\end{equation}

and

\begin{equation}
    \bar{S} =
    \frac{1}{B}
    \sum_{b=1}^{B}
    S_b^*
\end{equation}

denotes the mean number of reasoning steps across the batch.

The coefficient $\lambda = 0.01$ discourages unnecessary reasoning iterations by introducing a weak pressure toward efficient halting. In practice, this encourages the model to use additional reasoning depth only when it improves language modeling performance sufficiently to justify the extra computation.

\section{Experiments}
\label{sec:experiments}

\subsection{Training Details}

CosmicFish-HRM was trained on the 10B-token
\textit{CosmicSet} dataset for 27,500 iterations
using an effective batch size of 128 and a
context length of 512 tokens. The model achieved
a final validation loss of 3.36.

\begin{figure}[h]
    \centering
    \includegraphics[width=0.49\textwidth]{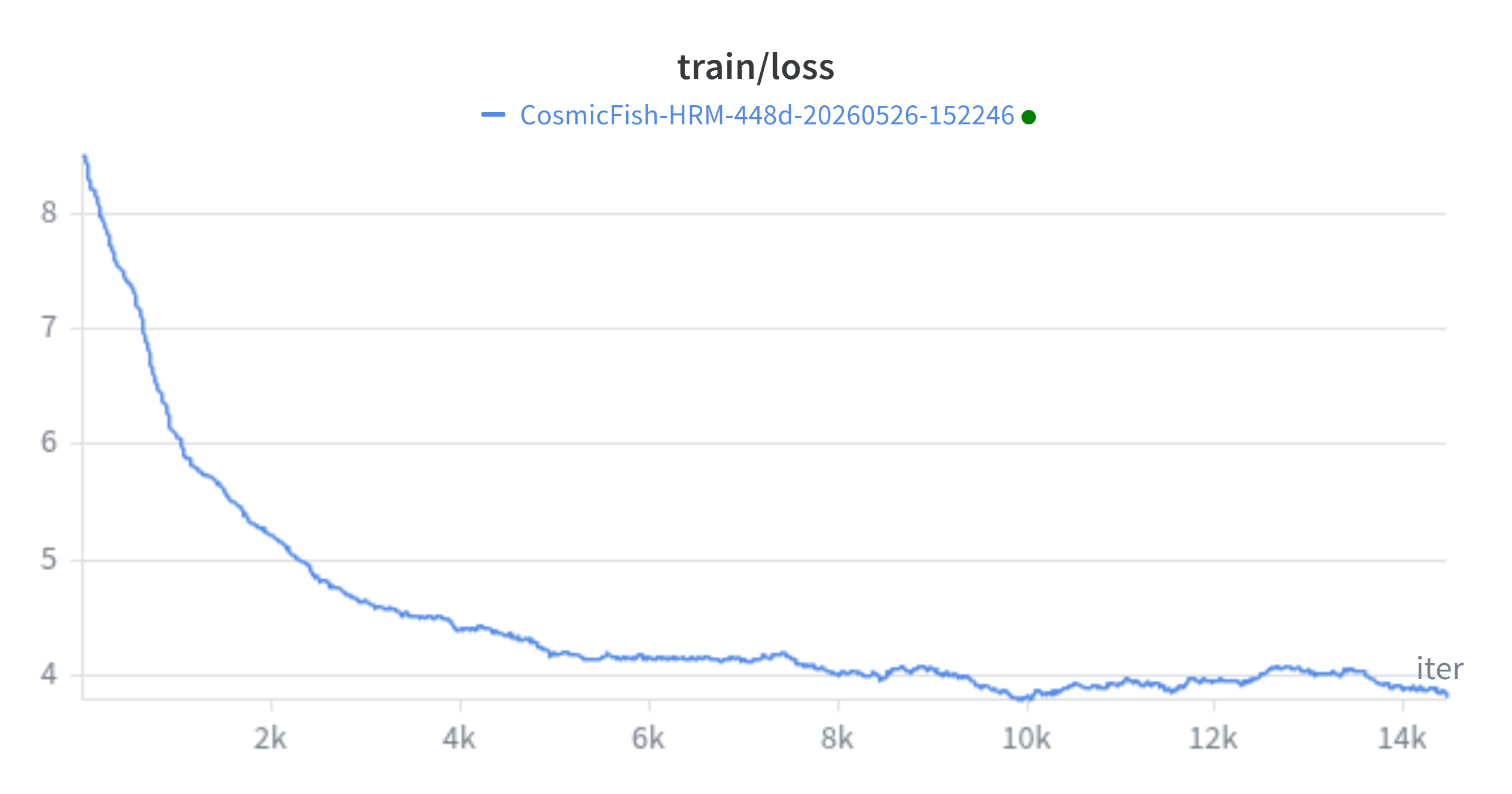}
    \includegraphics[width=0.49\textwidth]{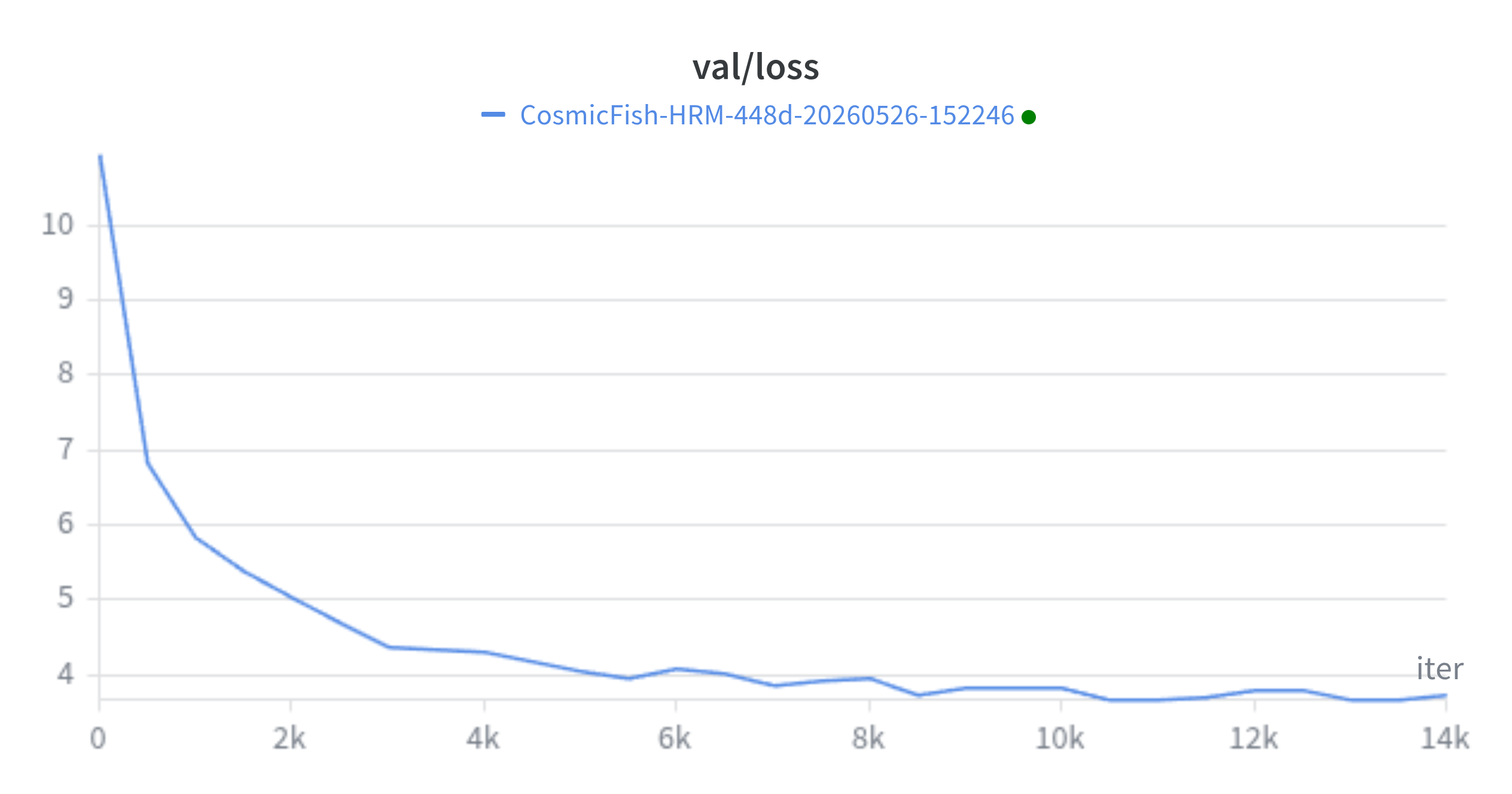}
    \caption{Training and validation loss curves for CosmicFish-HRM.}
    \label{fig:loss_curves}
\end{figure}

\begin{figure}[h]
    \centering
    \includegraphics[width=0.49\textwidth]{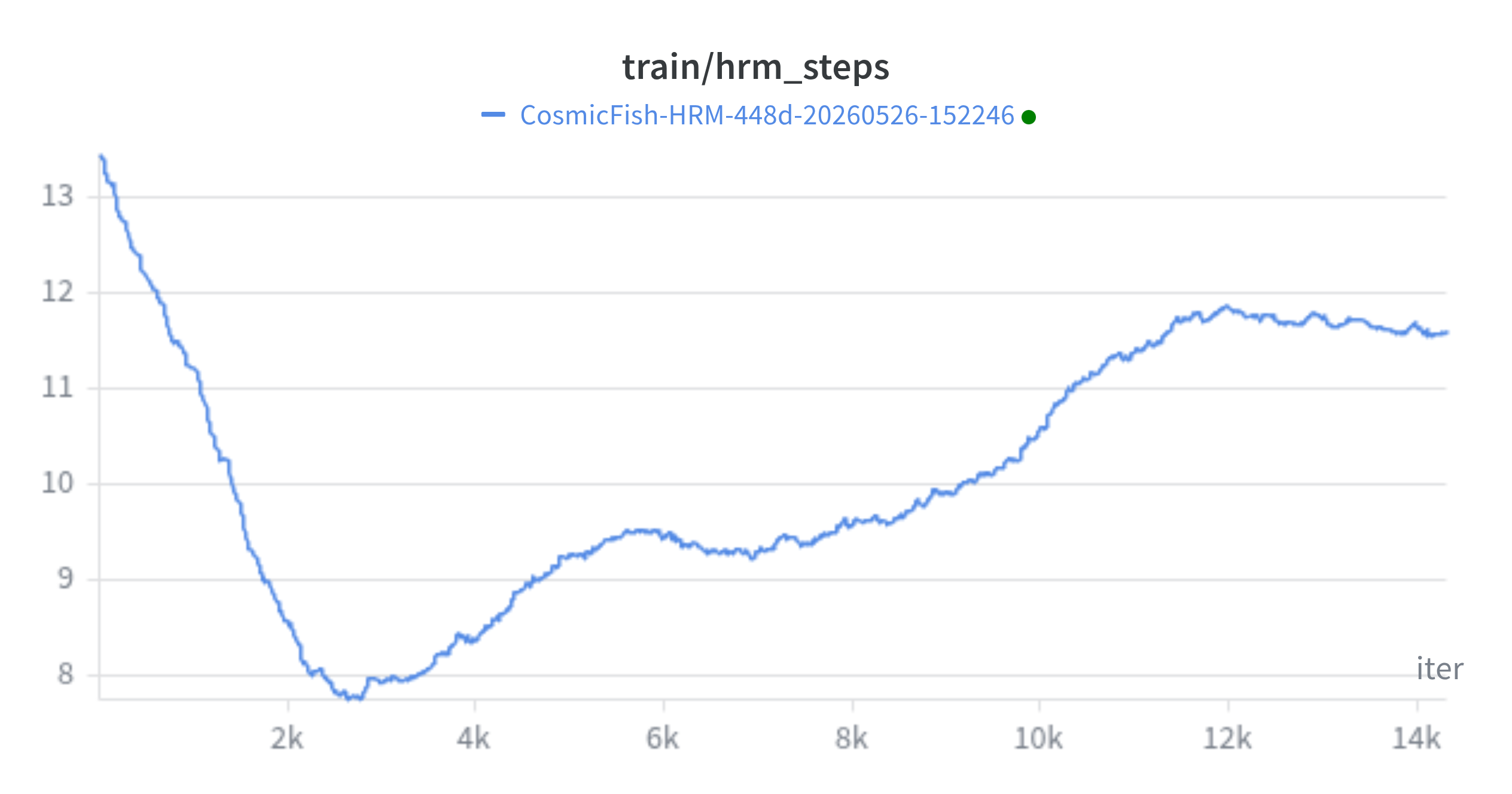}
    \includegraphics[width=0.49\textwidth]{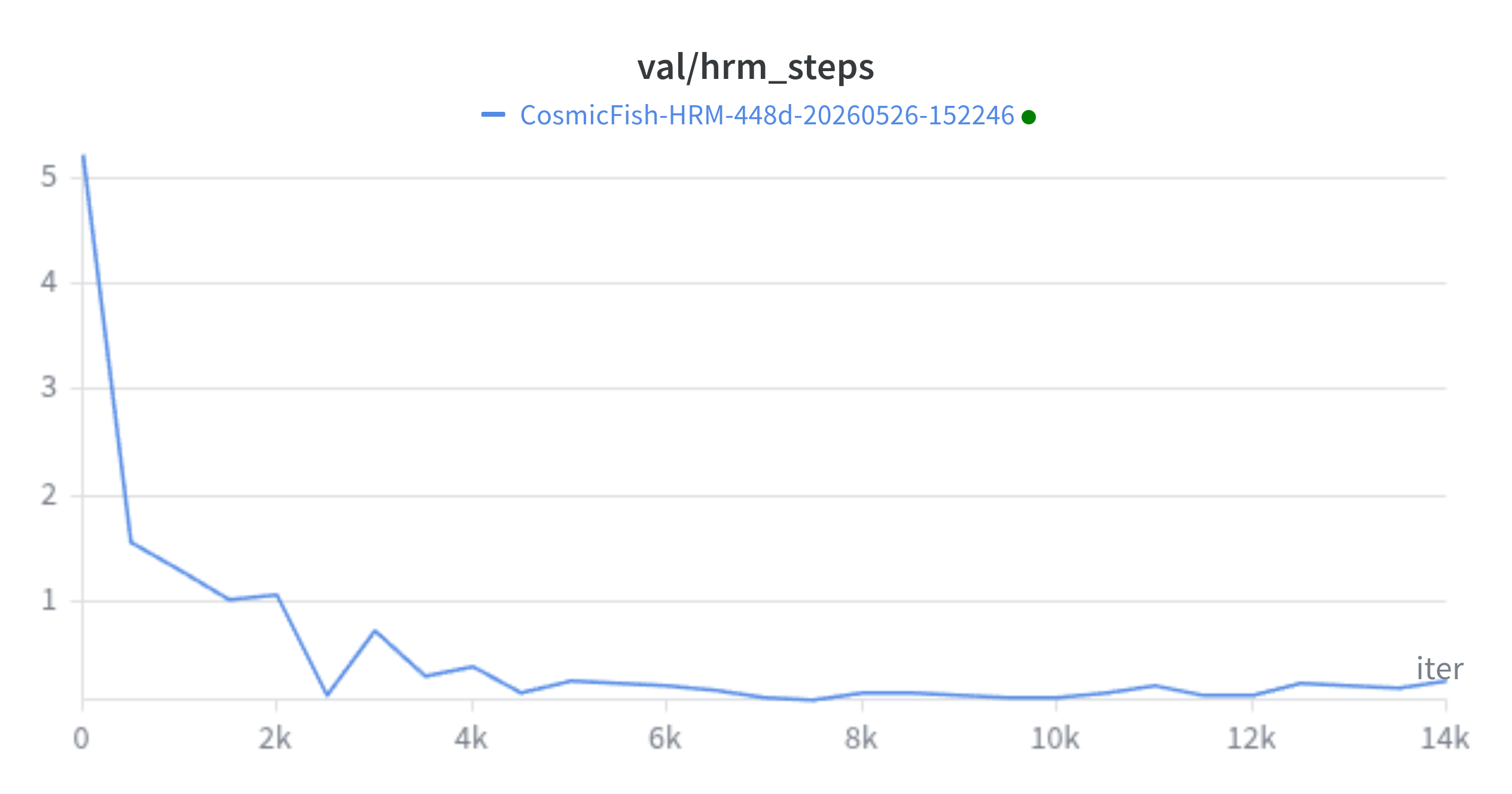}
    \caption{Average HRM reasoning steps during training and validation.}
    \label{fig:hrm_curves}
\end{figure}
\subsection{Setup}

We evaluate CosmicFish-HRM on a collection of standard language modeling and reasoning benchmarks including HellaSwag~\cite{zellers2019hellaswag}, PIQA~\cite{bisk2020piqa}, WinoGrande~\cite{sakaguchi2021winogrande}, TriviaQA~\cite{joshi2017triviaqa}, ARC-Easy~\cite{clark2018arc}, and Natural Questions~\cite{kwiatkowski2019nq}. These tasks span commonsense reasoning, physical reasoning, factual recall, coreference resolution, and open-domain question answering.

CosmicFish-HRM contains 82.77M parameters and is compared against earlier CosmicFish variants without the HRM reasoning core, as well as several publicly available transformer baselines of comparable scale including GPT-2 Small (117M)~\cite{radford2019gpt2}, OPT-125M~\cite{zhang2022opt}, and Pythia-160M~\cite{biderman2023pythia}. All evaluations are performed in the zero-shot setting.

It is worth noting that the benchmarks used here primarily measure general language modeling and shallow reasoning ability. They are not specifically designed to evaluate adaptive iterative reasoning behavior. As a result, these experiments should be interpreted primarily as an initial study of the architectural tradeoffs introduced by HRM-style reasoning inside compact language models rather than a definitive measurement of the architecture's long-horizon reasoning potential.

\subsection{Main Results}

\begin{table}[h]
\centering
\caption{Zero-shot benchmark evaluation. All scores are reported as accuracy (\%).}
\label{tab:benchmarks}
\resizebox{\textwidth}{!}{%
\begin{tabular}{lccc}
\toprule
\textbf{Model} & \textbf{HellaSwag} & \textbf{PIQA} & \textbf{WinoGrande} \\
\midrule
CosmicFish-90M  & 27.9 & 59.8 & 50.6 \\
CosmicFish-120M & 26.9 & 59.2 & 50.7 \\
\midrule
GPT-2 Small (117M) & 29.7 & 62.5 & 50.7 \\
OPT-125M & 30.6 & 62.6 & 52.9 \\
Pythia-160M & 29.4 & 62.1 & 52.8 \\
\midrule
\textbf{CosmicFish-HRM} & 26.2 & 58.1 & 50.7 \\
\bottomrule
\end{tabular}%
}
\end{table}

At compact scale, CosmicFish-HRM does not outperform conventional transformers of similar size on these benchmarks. We believe this reflects an important architectural tradeoff rather than a failure of adaptive reasoning itself. Unlike standard decoder-only transformers, CosmicFish-HRM allocates a portion of its parameter budget and computation toward recurrent reasoning infrastructure, halting behavior, and hierarchical state interaction. In a model with fewer than 100M parameters, this overhead represents a relatively large fraction of total capacity.

Our hypothesis is that this tradeoff changes as model size increases. The HRM reasoning infrastructure remains relatively small compared to the backbone transformer, meaning its proportional overhead decreases at larger scales while the adaptive reasoning behavior remains available. Under this view, compact-scale experiments primarily test whether adaptive reasoning behavior can emerge at all inside autoregressive language models, rather than whether such systems immediately dominate standard transformer baselines on shallow benchmark accuracy.

\subsection{Discussion}

Raw benchmark accuracy alone does not fully capture the behavior CosmicFish-HRM is designed to study. The central question is not simply whether the model achieves higher scores, but whether it learns to allocate reasoning depth non-uniformly across inputs.

Section~\ref{sec:analysis} examines this behavior directly by analyzing reasoning step allocation across tasks and prompts. In practice, the model tends to spend more reasoning iterations on inputs that appear structurally or semantically more difficult while halting early on simpler completions. This adaptive behavior represents the primary architectural contribution of the system and differentiates CosmicFish-HRM from fixed-depth transformer baselines.

We also note that many of the benchmarks used here emphasize short-context commonsense prediction rather than extended reasoning chains. Evaluating adaptive reasoning architectures on deeper reasoning-oriented tasks remains an important direction for future work.

\section{Analysis}
\label{sec:analysis}

The primary goal of CosmicFish-HRM is not simply to increase benchmark accuracy, but to investigate whether compact autoregressive language models can learn non-uniform reasoning behavior through adaptive computation. The central question, then, is whether the HRM core meaningfully changes how computation is allocated across inputs.

This section examines that behavior from three perspectives: reasoning step allocation across tasks, overall compute efficiency, and qualitative examples illustrating how the model responds to prompts of varying complexity.

\subsection{Adaptive Step Distribution}
\label{sec:analysis_steps}

An adaptive reasoning system should not behave like a fixed-depth transformer internally. If the halting mechanism is functioning meaningfully, more difficult inputs should tend to trigger longer reasoning trajectories, while simpler prompts should halt earlier.

To study this behavior, we measure the number of HRM reasoning iterations used across benchmark tasks.

\begin{figure}[h]
    \centering
    \includegraphics[width=0.85\linewidth]{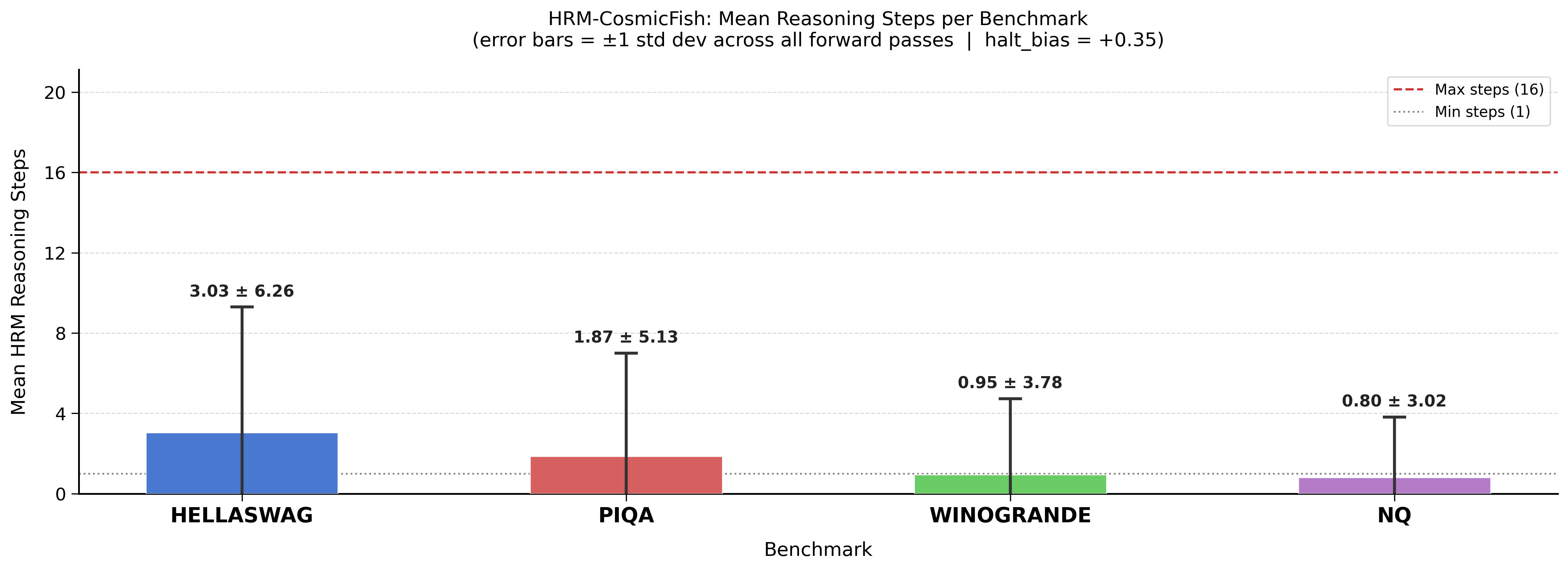}
    \caption{Mean HRM reasoning steps across benchmark tasks. More complex or retrieval-heavy tasks tend to trigger longer reasoning trajectories, while simpler prompts halt earlier.}
    \label{fig:step_dist}
\end{figure}

The resulting distributions suggest that the model does not collapse into a single fixed reasoning depth. Instead, reasoning trajectories vary substantially across tasks and prompts. Although this does not prove that the model has learned a complete notion of "reasoning difficulty," it does indicate that the halting mechanism is sensitive to differences in input structure and prediction uncertainty.

Importantly, this behavior emerges without manually designed stopping rules. The adaptive reasoning dynamics arise from the interaction between the learned halting head, recurrent reasoning states, and the lightweight step penalty introduced during training.

\subsection{Compute Efficiency}
\label{sec:analysis_efficiency}

An adaptive reasoning system is only useful in practice if it actually avoids unnecessary computation. Simply adding recurrent reasoning loops without meaningful halting would increase inference cost without providing adaptive behavior.

Table~\ref{tab:steps} reports the mean and standard deviation of reasoning steps across benchmark tasks relative to the maximum reasoning budget of 16 steps.

\begin{table}[h]
\centering
\caption{Mean HRM reasoning steps per benchmark task (maximum = 16). Lower values indicate earlier halting and reduced average computation.}
\vspace{5mm}
\label{tab:steps}
\begin{tabular}{lcc}
\toprule
\textbf{Benchmark} & \textbf{Mean Steps} & \textbf{Std. Dev.} \\
\midrule
HellaSwag & 3.033 & 6.263 \\
PIQA & 1.866 & 5.132 \\
WinoGrande & 0.954 & 3.777 \\
TriviaQA & 0.804 & 3.017 \\
\midrule
\textbf{Overall} & 2.681 & 5.949 \\
\bottomrule
\end{tabular}
\end{table}

The relatively low average reasoning depth suggests that the model frequently halts before reaching the maximum reasoning budget. At the same time, the high variance across samples indicates that reasoning depth remains highly input-dependent rather than collapsing into a near-constant policy.

We note, however, that these measurements should be interpreted cautiously. The current experiments are intended primarily as an initial behavioral analysis of adaptive reasoning dynamics at compact scale rather than a definitive compute-efficiency benchmark against optimized transformer inference systems.

\subsection{Qualitative Examples}
\label{sec:analysis_qualitative}

To make the adaptive reasoning behavior more concrete, Table~\ref{tab:examples} presents representative prompts alongside their average reasoning depth.

\begin{table*}[h]
\centering
\caption{Representative prompts and average reasoning depth. Simpler factual prompts tend to halt earlier, while prompts requiring abstraction or multi-step reasoning trigger deeper reasoning trajectories.}
\label{tab:examples}
\resizebox{\textwidth}{!}{%
\begin{tabular}{p{9cm} c}
\toprule
\textbf{Input Prompt} & \textbf{Mean Steps Used} \\
\midrule
\small\textit{``The capital of France is''} & 2.7805 \\[1.5ex]

\small\textit{``Photosynthesis is the process by which plants''} & 4.7722 \\[1.5ex]

\small\textit{``If all roses are flowers and some flowers fade quickly, what can we conclude about roses?''} & 7.0303 \\[1.5ex]

\small\textit{``A bat and a ball cost \$1.10 in total. The bat costs \$1.00 more than the ball. How much does the ball cost?''} & 8.4000 \\[1.5ex]

\bottomrule
\end{tabular}%
}
\end{table*}

The examples are intentionally chosen to span a rough spectrum of reasoning difficulty. Short factual completions generally require relatively few reasoning iterations, while prompts involving abstraction, chained inference, or cognitive reflection tend to trigger deeper reasoning trajectories.

The bat-and-ball example is particularly interesting because it is widely used in cognitive psychology as a probe distinguishing fast intuitive responses from slower deliberate reasoning~\cite{kahneman2011thinking}. Seeing the model allocate substantially more reasoning steps to this prompt than to simple factual recall suggests that the adaptive halting mechanism is responding to meaningful differences in prompt structure rather than behaving randomly.

That said, these examples should not be interpreted as evidence of human-like reasoning. The current results are better understood as evidence that the model has learned variable-depth computational behavior conditioned on input characteristics.

\subsection{Summary}
\label{sec:analysis_summary}

Taken together, the step distributions and qualitative examples suggest that CosmicFish-HRM learns non-uniform reasoning behavior rather than relying on a fixed computation budget for every prompt. The model allocates different reasoning depths across tasks and inputs, often halting early for simple completions while extending computation for prompts that appear structurally more difficult.

At this scale, the adaptive reasoning infrastructure introduces clear tradeoffs in raw benchmark performance. Even so, the experiments provide evidence that hierarchical recurrent reasoning can emerge inside compact autoregressive language models without hand-crafted reasoning schedules or explicit reasoning supervision. We view this as an initial step toward studying adaptive reasoning depth as an architectural property of future language models rather than treating reasoning purely as a function of parameter scale.

\section{Conclusion}
\label{sec:conclusion}

We introduced \textbf{CosmicFish-HRM}, a compact 82.77M parameter language model designed to explore adaptive reasoning depth inside autoregressive transformers. By integrating a Hierarchical Reasoning Module between transformer stacks, the model is able to allocate different amounts of computation across inputs rather than applying a fixed reasoning depth uniformly during inference.

The experiments presented in this work suggest that compact language models can learn variable-depth computational behavior through recurrent hierarchical reasoning and learned adaptive halting. Although CosmicFish-HRM does not outperform conventional transformers of similar size on standard benchmark accuracy, the results indicate that meaningful adaptive reasoning dynamics can emerge even at relatively small scales.

At the same time, the work highlights an important tradeoff. In compact models, the reasoning infrastructure itself occupies a non-trivial fraction of the total parameter budget and compute capacity. This creates tension between adaptive reasoning flexibility and raw language modeling performance. We believe this tradeoff may become substantially more favorable at larger scales, where the relative overhead of the HRM core decreases while the adaptive computation behavior remains available.

More broadly, this work explores a different perspective on efficiency in language models. Much of the field has focused on making large models smaller through compression techniques such as pruning, quantization, and distillation. CosmicFish-HRM instead investigates whether smaller models can become more capable by using computation more selectively and adaptively during inference.

We do not view the current system as a final solution or a replacement for conventional transformer scaling. Rather, we view it as an initial architectural exploration into adaptive reasoning depth as a property of language models. The results suggest that reasoning behavior may depend not only on parameter count, but also on how computation is distributed across inputs and reasoning trajectories.

\subsection*{Future Work}

Several directions seem particularly important moving forward.

\paragraph{Scaling Behavior.}

The most immediate question is how HRM-style reasoning behaves at larger scales. In larger models, the relative overhead of the recurrent reasoning infrastructure becomes much smaller compared to the backbone transformer. Understanding whether adaptive reasoning becomes more effective under those conditions remains an open problem.

\paragraph{Reasoning-Oriented Evaluation.}

Many existing language model benchmarks primarily evaluate short-context prediction and shallow commonsense reasoning. Future work should evaluate adaptive reasoning architectures on tasks requiring longer reasoning chains, planning, abstraction, mathematical reasoning, and multi-hop retrieval.

\paragraph{Mixture-of-Experts Integration.}

Combining adaptive reasoning depth with sparse expert routing~\cite{shazeer2017moe} could produce architectures capable of scaling both computation depth and representational capacity efficiently. An HRM system operating over sparse expert modules may provide a more scalable form of adaptive inference.

\paragraph{Improved HRM Integration.}

The current architecture treats the HRM core as a relatively modular reasoning stage inserted between transformer stacks. Future systems may benefit from tighter interaction between transformer representations and recurrent reasoning states, including cross-attention between HRM states and intermediate transformer activations.

\paragraph{Interpretability and Internal Dynamics.}

The internal behavior of the high-level and low-level reasoning states remains poorly understood. Future work should investigate how representations evolve across reasoning iterations, whether distinct reasoning patterns emerge at different depths, and how the adaptive halting mechanism relates to uncertainty or prediction entropy.

\paragraph{Instruction Tuning and Alignment.}

Instruction tuning and preference optimization~\cite{ouyang2022rlhf} may also interact in interesting ways with adaptive reasoning depth. It remains unclear whether aligned or instruction-following models learn different halting behaviors from purely pretrained systems.

\section*{Code Availability}

The implementation of CosmicFish-HRM and related
resources are publicly available at:

\url{https://github.com/MistyozAI/CosmicFish-HRM}

\appendix
\section*{Appendix}

\section{Qualitative Examples}
\label{sec:appendix_examples}

Representative generations from the instruction-tuned version of CosmicFish-HRM together with HRM reasoning statistics.

\subsection{Greeting}

\noindent
\textbf{Prompt:}
\begin{verbatim}
Hello
\end{verbatim}

\noindent
\textbf{Response:}
\begin{verbatim}
Hey! How's everything going?
What brings you here today?
\end{verbatim}

\noindent
\textbf{Avg. HRM steps/token:} 15.00

\vspace{2mm}

\subsection{Identity Question}

\noindent
\textbf{Prompt:}
\begin{verbatim}
Who are you?
\end{verbatim}

\noindent
\textbf{Response:}
\begin{verbatim}
Hey! I'm CosmicFish, an AI assistant built
by Mistyoz AI in Hyderabad.
\end{verbatim}

\noindent
\textbf{Avg. HRM steps/token:} 13.59

\vspace{2mm}

\subsection{Poem Generation}

\noindent
\textbf{Prompt:}
\begin{verbatim}
Write a poem
\end{verbatim}

\noindent
\textbf{Response:}

\begin{verbatim}
The wind howled in the breeze,
A symphony of life,
As we wander through the night.
\end{verbatim}

\noindent
\textbf{Avg. HRM steps/token:} 8.66

\vspace{2mm}

\subsection{Definition Generation}

\noindent
\textbf{Prompt:}
\begin{verbatim}
Define AI
\end{verbatim}

\noindent
\textbf{Response:}

\begin{verbatim}
Artificial Intelligence (AI) refers to the
simulation of human intelligence in machines.
\end{verbatim}

\noindent
\textbf{Avg. HRM steps/token:} 6.30

\vspace{2mm}

\subsection{Arithmetic}

\noindent
\textbf{Prompt:}
\begin{verbatim}
What is 1 + 2?
\end{verbatim}

\noindent
\textbf{Response:}

\begin{verbatim}
1 + 2 = 3
\end{verbatim}

\noindent
\textbf{Avg. HRM steps/token:} 5.79

\vspace{2mm}

\subsection{Code Generation}

\noindent
\textbf{Prompt:}
\begin{verbatim}
Write a python program to add two numbers
\end{verbatim}

\noindent
\textbf{Response:}

\begin{verbatim}
def add_numbers(num1, num2):
    return num1 + num2
\end{verbatim}

\noindent
\textbf{Avg. HRM steps/token:} 10.01

\section{Failure Cases}
\label{sec:appendix_failures}

\subsection{Factual Hallucination}

\noindent
\textbf{Prompt:}
\begin{verbatim}
What is the capital of India?
\end{verbatim}

\noindent
\textbf{Response:}

\begin{verbatim}
India has an estimated population of around
1.5 billion people.
\end{verbatim}

\noindent
\textbf{Avg. HRM steps/token:} 10.46

\vspace{2mm}

\subsection{Arithmetic Failure}

\noindent
\textbf{Prompt:}
\begin{verbatim}
What is 5 + 5?
\end{verbatim}

\noindent
\textbf{Response:}

\begin{verbatim}
5 + 5 = 25
\end{verbatim}

\noindent
\textbf{Avg. HRM steps/token:} 10.54

\section{Summary}
\label{sec:appendix_summary}

These examples illustrate that CosmicFish-HRM exhibits non-uniform reasoning behavior across prompts and tasks. Different inputs trigger different reasoning depths, though deeper reasoning does not necessarily guarantee correctness.

{\small
\bibliographystyle{plainnat}
\bibliography{references}
}

\end{document}